\documentclass[10pt,twocolumn,letterpaper]{article}

\usepackage{iccv}
\usepackage{times}
\usepackage{epsfig}
\usepackage{graphicx}
\usepackage{amsmath}
\usepackage{amssymb}

\usepackage{multirow}
\usepackage{booktabs}
\usepackage{subcaption}
\usepackage{arydshln}

\DeclareMathOperator*{\argmax}{arg\,max}

\usepackage[pagebackref=true,breaklinks=true,letterpaper=true,colorlinks,bookmarks=false]{hyperref}

\usepackage[capitalize]{cleveref}
\crefname{section}{Sec.}{Secs.}
\Crefname{section}{Section}{Sections}
\Crefname{table}{Table}{Tables}
\crefname{table}{Tab.}{Tabs.}

\iccvfinalcopy 

\def\iccvPaperID{****} 

\ificcvfinal\fi

\begin{document}

\title{Distilled Reverse Attention Network for Open-world Compositional Zero-Shot Learning}

\author{Yun Li\\
University of New South Wales\\
Institution1 address\\
{\tt\small yun.li5@unsw.edu.au}
\and
Zhe Liu\\
Jiangnan University\\
{\tt\small zheliu912@gmail.com}
\and
Saurav Jha\\
University of New South Wales\\
{\tt\small saurav.jha@unsw.edu.au}
\and
Sally Cripps\\
University of Technology Sydney\\
{\tt\small sally.cripps@uts.edu.au}
\and
Lina Yao\\
University of New South Wales\\
{\tt\small lina.yao@unsw.edu.au}
}

\maketitle
\ificcvfinal\thispagestyle{empty}\fi

\begin{abstract}
   Open-World Compositional Zero-Shot Learning (OW-CZSL) aims to recognize new compositions of seen attributes and objects. In OW-CZSL, methods built on the conventional closed-world setting degrade severely due to the unconstrained OW test space. While previous works alleviate the issue by pruning compositions according to external knowledge or correlations in seen pairs, they introduce biases that harm the generalization. Some methods thus predict state and object with independently constructed and trained classifiers, ignoring that attributes are highly context-dependent and visually entangled with objects. In this paper, we propose a novel Distilled Reverse Attention Network to address the challenges. We also model attributes and objects separately but with different motivations, capturing contextuality and locality, respectively. We further design a reverse-and-distill strategy that learns disentangled representations of elementary components in training data supervised by reverse attention and knowledge distillation. We conduct experiments on three datasets and consistently achieve state-of-the-art (SOTA) performance.
\end{abstract}

\section{Introduction}
\label{sec:intro}

Humans can recognize complex concepts never seen before (\eg, the pink elephant) by composing their knowledge of familiar visual primitives (elephants and other pink objects). This ability of compositional learning is considered a hallmark of human intelligence~\cite{lake2014towards} that deep learning methods clearly lack~\cite{lake2017building}. Deep learning often requires a large quantity of labeled examples to train. However, real-world instances follow a long-tailed distribution~\cite{van2017devil,wang2017learning}, making it impractical to gather supervision for all categories. Compositional Zero-shot Learning (CZSL) mimics the human ability to tackle these issues~\cite{purushwalkam2019task,li2020symmetry,naeem2021learning,karthik2022kg}. 

\begin{figure}
\small
\centering
  \includegraphics[width=0.85\linewidth]{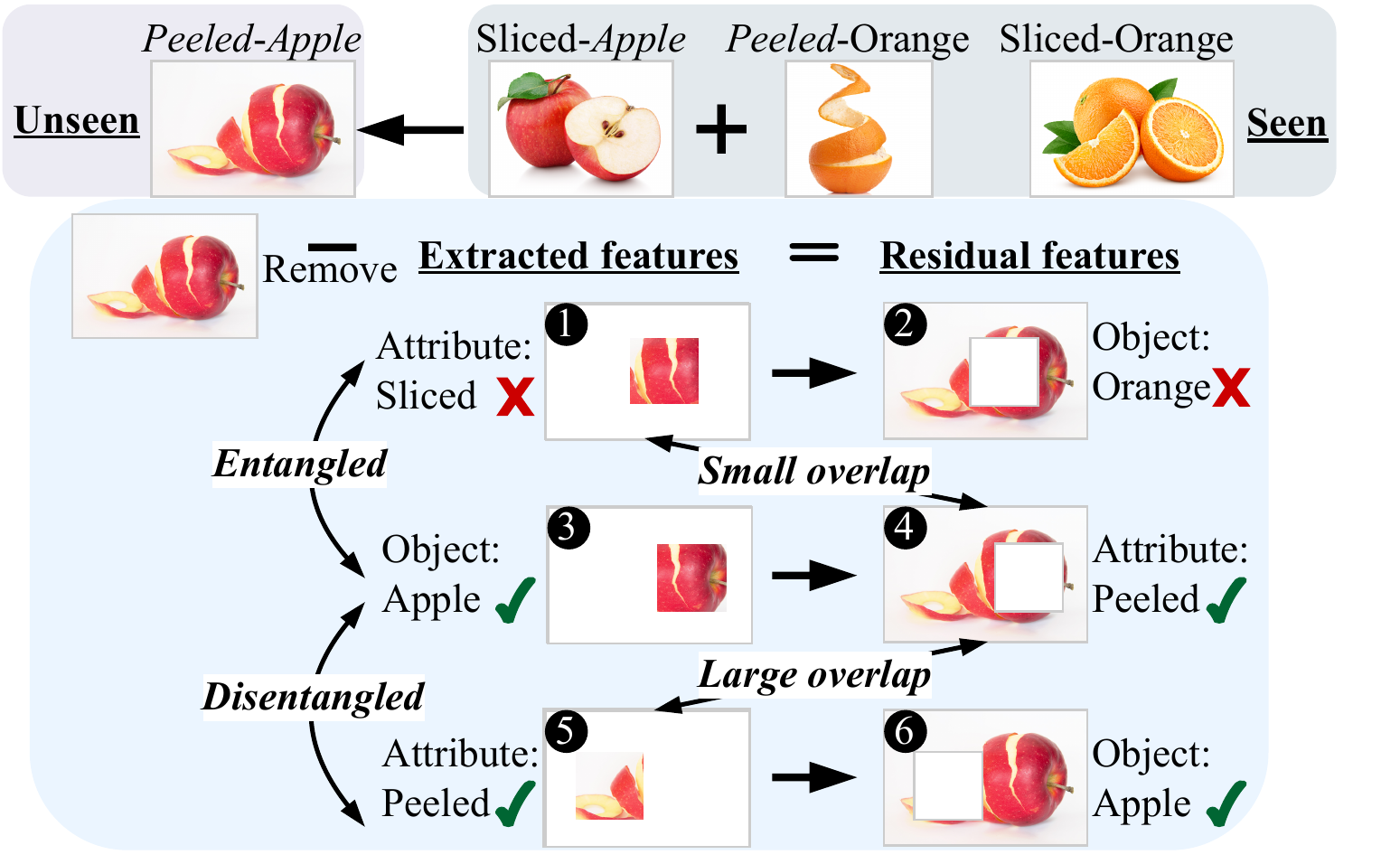}
  \caption{Motivation behind our disentangling strategy for OW-CZSL. When extracted features of objects and attributes are disentangled (images 3 and 5), their residual features (images 4 and 6) carry sufficient information about each other to classify correctly, and produce large overlap between the object residuals and the attribute features (images 4 and 5). For entangled attribute-object features (images 1 and 3), the phenomena are otherwise reversed (image 2: few object information; images 1 and 4: small overlap).}
    \label{fig:intro}
\end{figure}

CZSL learns the compositionality of seen objects (\eg fruits, animals, etc.) and attributes (\eg colors, sizes, etc.) as primitives to recognize unseen attribute-object pairs. For example, CZSL composes and generalizes \textit{Peeled}-Orange and Sliced-\textit{Apple} to \textit{Peeled-Apple} (\cref{fig:intro}).
Conventional CZSL methods characterize closed-world (CW-CZSL) settings~\cite{purushwalkam2019task,misra2017red,nagarajan2018attributes,li2020symmetry}, where unseen attribute-object pairs contained in test images are given as priors to restrict the search space. For example, the test space of the widely-used benchmark MIT-States~\cite{isola2015discovering} is simplified to 1,662 compositions out of 28,175 possible pairs (115 attributes $\times$ 245 objects) for CW-CZSL. This setup fundamentally reduces the generalization ability of CZSL models. Therefore, in this work, we study a more realistic and challenging task: unconstrained Open-World CZSL (OW-CZSL)~\cite{karthik2021revisiting,karthik2022kg,mancini2021open,mancini2022learning}, where arbitrary compositions may appear at test time.

A notable line of works for CW-CZSL projects attribute-object pairs and images onto a shared embedding space to perform similarity-based composition classification ~\cite{xu2021zero,naeem2021learning,wei2019adversarial}. However, their performances severely degrade for OW-CZSL ~\cite{mancini2021open} due to greatly expanded output space (\eg, $\sim$ 17 times in MIT-States). Thus, some works adapt them to OW-CZSL by pruning OW composition space based on feasibility scores calculated according to linguistic side information~\cite{karthik2022kg} or seen attribute-object dependencies~\cite{mancini2021open,mancini2022learning}. Such scores inevitably introduce biases caused by distribution shifts between images and external linguistic knowledge bases  or seen and unseen compositions, resulting in visual-semantic inconsistent or seen-biased predictions. Therefore, for OW-CZSL, we follow another direction that adopts two parallel discriminative modules to infer objects and attributes respectively, reducing composition search to separate attribute and object search~\cite{karthik2022kg,karthik2021revisiting,li2022siamese,xu2021relation}.

Despite the success of separate modeling techniques in CW- and OW-CZSL, these ignore the intrinsic differences between attributes and objects~\cite{li2022siamese,xu2021relation,karthik2022kg,karthik2021revisiting}. Children, for instance, learn nouns faster than adjectives because they relate to context differently~\cite{gasser1998learning}. Similarly, visual primitives of attributes (often adjectives) are more context-sensitive than objects (usually nouns)~\cite{misra2017red,nagarajan2018attributes}. For example, \textit{Small} in Small-Cat and Small-Building is not visually equivalent, while \textit{Tomato} in Red-Tomato and Fresh-Tomato is similar. Extracting attribute and object features using identical structures~\cite{karthik2022kg,karthik2021revisiting} without considering the heavier context dependencies of attributes may impair the discrimination.

Another bottleneck for separate modeling is visual entanglement. Taking \cref{fig:intro} as an example, given an image of the unseen composition, \textit{i.e.,} Peeled-Apple, it is hard to distinguish which visual features are Apple and which ones are Peeled.
The extracted features of attributes and objects are highly entangled (images 1 and 3), leading to a wrong prediction biased towards the seen pairs, \textit{i.e.,} Sliced-Apple. 
Some efforts disentangle the embeddings in CW-CZSL~\cite{saini2022disentangling,atzmon2020causal,xu2021relation,li2022siamese}. However, they either learn pair-wise attribute-object correlations in compositional space~\cite{ruis2021independent,atzmon2020causal} or adopt generative methods to synthesize samples for all pairs~\cite{saini2022disentangling,li2022siamese}, thus making them infeasible for OW-CZSL due to the drastically expanded output space.

To address these issues, we propose the Distilled Reverse Attention Network (DRANet) that extracts and disentangles visual primitives of attributes and objects for OW-CZSL.
First, we design attribute/object-specific networks to extract their features differently according to their characteristics. As suggested by ~\cite{8578911}, Convolutional Neural Networks (CNNs), used to extract visual embeddings in CZSL, are built on top of local neighborhoods and thus cannot capture long-range context. Therefore, we adapt non-local attention blocks~\cite{8578911,DBLP:conf/cvpr/FuLT0BFL19} to model spatial and channel contextual relationships for attribute learning while adopting local attention to focus on essential parts for object recognition.

Second, we design an attention-based disentangling strategy for OW-CZSL, namely \textit{Reverse-and-Distill}.
This strategy is based on the observation that humans can still recognize \textit{Apple} after removing \textit{Peeled} from the images of Peeled-Apple.
Intuitively, if learned primitives of attributes and objects are disentangled, removing either of them from the feature space will not affect the classification of the other.
Thus, object predictions after erasing the attribute features (or attribute predictions after object removal) can indicate unraveling degree of attribute and object features. 
For example, as shown in \cref{fig:intro}, models can still recognize Apple (image 6) after removing attribute features when primitives are disentangled (images 3 and 5) but fail (image 2) when entangled (images 1 and 3).
Given that feature removal is intractable in practice, we approximate it by reversing attention. We then achieve attribute and object feature disentanglement by supervising their residuals to crossly carry sufficient object and attribute information.
Besides, when attribute and object features are disentangled, the overlaps between attribute features and object residuals (or object features and attribute residuals) become large (seeing images 4 and 5 or 3 and 6 in \cref{fig:intro}). We enlarge such overlaps by distilling primitives to learn from mutual residuals for further unraveling.

In summary, our contributions are as follows: 
1) We propose the DRANet for OW-CZSL. DRANet employs distinct extractors to capture attribute and object features, enhancing contextuality and locality, simultaneously. 
2) We design the reverse-and-distill strategy to disentangle the attribute and object embeddings in OW-CZSL, where existing disentangling methods in CW-CZSL are impractical.
3) We achieve SOTA performance on three benchmark datasets, and analyze the limitations and extensibility of our model.

\section{Related work}

\textbf{Compositional Zero-Shot Learning (CZSL)} aims to recognize unseen concepts by composing learned attribute and object primitives. A typical schema of CZSL is to learn joint representations of compositions~\cite{wang2019task,naeem2021learning,ruis2021independent,wei2019adversarial,xu2021zero}. \cite{naeem2021learning} establishes element and composition relationships in a graph space. \cite{purushwalkam2019task} uses a gating network to generate a unified classifier for compositions. ~\cite{yang2020learning} refines composition embeddings by hierarchically constructing concepts. 
Other methods try to model attributes as transformations applied to objects~\cite{li2020symmetry,nagarajan2018attributes} and learn a classifier based on objects modified by attributes. The transformation can be linear projection~\cite{nagarajan2018attributes} or symmetry coupling and decoupling~\cite{li2020symmetry}. 

\begin{figure*}
\small
\centering
  \includegraphics[width=0.8\linewidth]{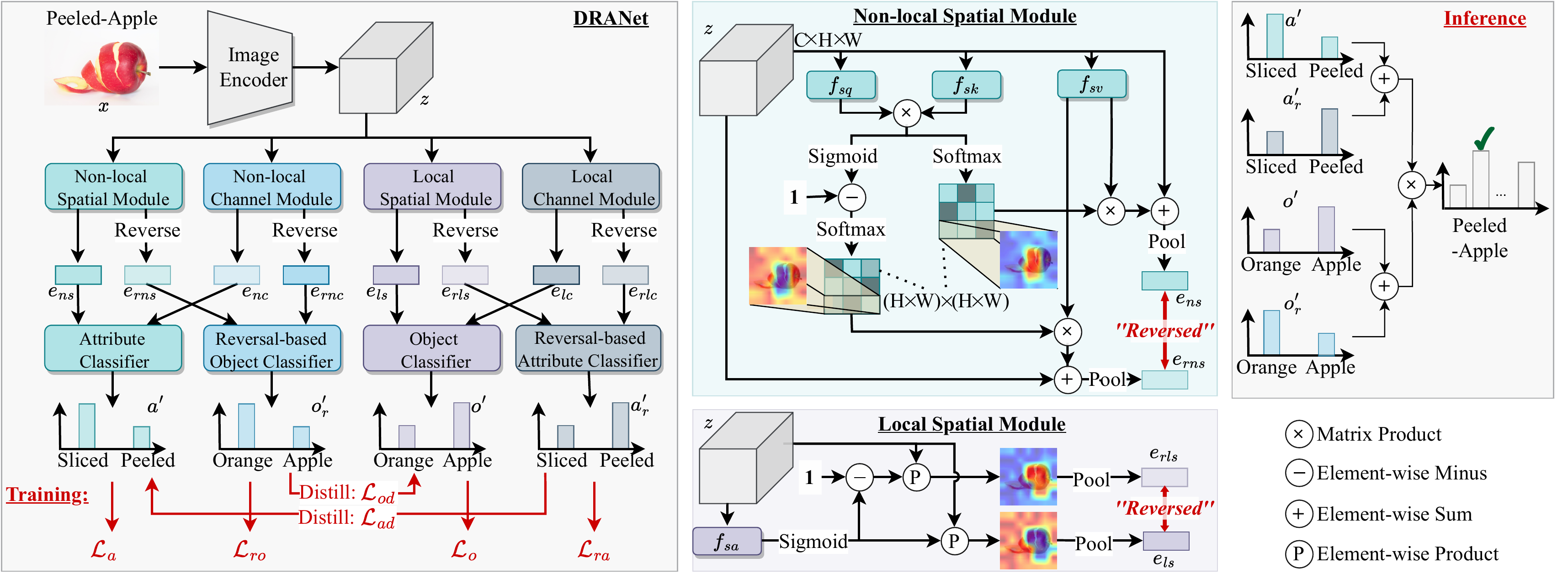}
    \caption{\textbf{DRANet} Overview.  It contains four modules to extract non-local and local features from spatial and channel dimensions. The concatenated spatial-channel embeddings from the non-local and local modules are used to predict attributes and objects, respectively. Their reversed knowledge is swapped as inputs for reversal-based object and attribute classifiers, respectively. The model is optimized with four classification losses and reversal-oriented distillation losses. The \textbf{Non-local} and \textbf{Local Spatial Modules} are based on non-local attention ~\cite{8578911} and soft attention~\cite{xie2019attentive}, respectively, and adapted using reverse attention for attribute-object disentanglement. During \textbf{inference}, all the results are combined for final predictions.}
    \label{fig:model}
\end{figure*}

Another mainstream methods model attributes and objects separately~\cite{xu2021relation,li2022siamese,saini2022disentangling,atzmon2020causal} to reduce composition learning into attribute and object learning. \cite{xu2021relation} employs a block memory network to generate features for concepts.  \cite{li2022siamese,saini2022disentangling} compare images with the same objects or objects to decompose visual primitives. Among them, some works~\cite{li2022siamese,saini2022disentangling,atzmon2020causal} find that isolated modeling ignores attribute-object interactions and thus proposes to \textbf{disentangle attributes and objects} for CW-CZSL with affinity estimation~\cite{saini2022disentangling}, contrastive learning~\cite{li2022siamese}, or cutting the confounding links~\cite{atzmon2020causal}. In this work, we design a new disentangling strategy suitable for OW-CZSL, \textit{i.e.}, reverse-and-distill. It takes a single image as input without image pair comparisons or sample generations~\cite{li2022siamese,saini2022disentangling}, regularizes and distills feature extraction via reverse attention to unravel attributes and objects.

\textbf{Open-world CZSL} (OW-CZSL) is more challenging due to its relaxed constraints on the output space~\cite{mancini2021open,karthik2021revisiting,karthik2022kg,mancini2022learning}. Feasibility~\cite{mancini2021open} is estimated to remove compositions by using ConceptNet to measure attribute-object compatibility~\cite{karthik2022kg}, or constructing graph convolutional networks to model primitive correlations~\cite{mancini2022learning}. As in~\cite{karthik2022kg,karthik2021revisiting}, we predict objects and attributes separately, different in that their predictions are in isolation, while we, as the first disentangling attempt in OW-CZSL, untwine two branches mutually and collaboratively for better generalization.

\textbf{Attention mechanisms} are commonly adopted in computer vision tasks such as scene segmentation~\cite{fu2020scene,DBLP:conf/cvpr/FuLT0BFL19}, image classification~\cite{chen2021crossvit,dong2022weighted}, or Zero-Shot Learning (ZSL)~\cite{li2022entropy,li2022balancing} that closely relates to CZSL. In ZSL, attention mechanisms are usually used to capture subtle visual differences~\cite{li2022balancing} or locate semantics-rich regions to improve attribute-visual compatibility~\cite{li2022entropy,liu2021rethink}.
Despite the success of attention mechanisms in vision tasks, as most CZSL tasks focus on how to explore compositional nature rather than visual representation learning, incorporation of visual attention in CZSL is underexplored. A previous work~\cite{xu2021relation} in CZSL adopts attention, but in the linguistic view. In this paper, we utilize attention in visual cues, adopting non-local attention inspired by~\cite{wang2018non,DBLP:conf/cvpr/FuLT0BFL19} to capture contextuality and using local attention to enhance visual distinction. With visual attention, DRANet can extract context without external linguistic knowledge (\eg, pre-trained word embeddings~\cite{nagarajan2018attributes}).

\section{Method}

\textbf{Problem definitions and notations.} CZSL models images as compositions of attributes $a\in\mathcal{A}$ and objects $o\in\mathcal{O}$. Suppose $\mathcal{A}$, $\mathcal{O}$, and training data $\mathcal{S}=\{(x, y)|x\in{X^{S}},y\in{Y^{S}}\}$ from seen compositions (compositions with labeled samples) are given, where $x\in{X^{S}}$ is an image with label $y\in{Y^{S}}$; $y$ is a tuple $(a,o)$ of attribute-object labels and $a\in\mathcal{A},o\in\mathcal{O}$. Given a test set $\mathcal{T}=\{(x, y)|x\in{X^{\mathcal{T}}},y\in{Y^{\mathcal{T}}}\}$, CZSL aims to predict the label $y\in{Y^{\mathcal{T}}}$ for each image $x\in{X^{\mathcal{T}}}$. 
For OW-CZSL,  $Y^{\mathcal{T}}$ is the set of all possible attribute-object pairs $Y^{\mathcal{T}}=\mathcal{A}\times\mathcal{O}$.
More specifically,  output space in OW-CZSL consists of seen compositions $Y^{\mathcal{S}}$, unseen compositions $Y^{\mathcal{U}}$ without training samples, and pairs not present in the dataset. Note that seen and unseen compositions are disjoint, i.e., $Y^{\mathcal{S}}\cap Y^{\mathcal{U}}=\emptyset$. To bridge them, all attributes and objects in $Y^{\mathcal{U}}$ appear as label elements in $Y^{\mathcal{S}}$, \textit{i.e.}, seen elements form unseen pairs.

\textbf{Overview.} As shown in \cref{fig:model}, our DRANet includes non-local and local modules. Under the constraints of attribute and object classification losses, non-local blocks attempt to extract spatial and channel contextuality for attribute learning; local blocks aim to discover important regions and channels for object recognition. Attention-based reversing operations mimic feature erasures to encourage the attribute-object disentanglement supervised by the reversal-based classification losses. Distillation losses further encourage mutually exclusive learning of non-local and local blocks throughout the training process.

\subsection{Non-local Networks for Attributes.} Contextuality is crucial for attribute understanding~\cite{misra2017red,nagarajan2018attributes} due to its heavy dependency on context. Thus for attributes, we adapt non-local attention~\cite{8578911,DBLP:conf/cvpr/FuLT0BFL19} to relate high-response regions and channels with themselves and with externals to capture contextuality.
However, the extracted attribute features may be highly entangled with object features; thus, we design the reverse attention mechanism and incorporate it with the non-local blocks to perform feature disentanglement. In this section, we introduce the design of reverse attention in attribute learning. Object reverse attention, and how to realize the decoupling by the reverse attention are detailed in \cref{sec:obj,sec:dis}, respectively.

\textbf{Non-local Spatial Module (NSM).}
Given an image $x$, the image encoder embeds $x$ to obtain the feature map $z\in \mathbb{R}^{C\times H\times W}$. Then, as shown in \cref{fig:model} (left and top-center in the figure), NSM feeds $z$ to three one-layer $1\times 1$ CNNs, i.e., $f_{sq}$, $f_{sk}$, and $f_{sv}$, to generate the query, key, and value maps ($e_{sq}$, $e_{sk}$, and $e_{sv}$, respectively), where $\{e_{sq}, e_{sk}\}\in\mathbb{R}^{c\times H\times W}$ ($c$ is a reduced channel number to save computations), and $e_{sv}\in\mathbb{R}^{C\times H\times W}$. 
We reshape $e_{sq}$ and $e_{sv}$ to $\mathbb{R}^{c\times N}$, where $N=H\times W$, and perform a dot product between the transpose of $e_{sq}$ and $e_{sk}$: $w_{s}=e_{sq}^Te_{sk}$. 

To capture the contextuality, we then normalize $w_{s}$  with $\mathit{Softmax}$ to calculate the non-local attention map and multiply the reshaped $e_{sv}\in\mathbb{R}^{C\times N}$ with the transpose of the attention map. We then construct residual connections by adding the product (reshaped to $\mathbb{R}^{C\times H \times W}$) to $x$, and pool the sum to obtain the final non-local spatial outputs $e_{ns}$: 
\begin{equation}
\small
    e_{ns} = \mathit{Pool}(\alpha e_{sv}\mathit{Softmax}(w_{s})^T + x)
\end{equation}
where $\alpha$ is a learnable scale factor that is initialized to zero and is gradually optimized, and $Pool()$ is the average pooling function. For each position, $e_{ns}$ computes a weighted sum of the features across all positions and the original features $x$, contributing to a global contextual view, thus improving the attribute representation learning.

We also calculate the reversed embeddings based on $w_s$. We first use $\mathit{Sigmoid}$ to activate $w_s$ into (0,1) and subtract it from 1 to reverse the focus. We then apply a $\mathit{Softmax}$ layer to generate the reversed attention and calculate the reversed non-local spatial embedding $e_{rns}$: 
\begin{equation}
\small
    e_{rns} = \mathit{Pool}(\alpha e_{sv}\mathit{Softmax}(1-\mathit{Sigmoid}(w_{s}))^T + x)
\end{equation}

For the non-local spatial attention maps, the overall size is $N\times N$, \textit{i.e.,} $(H\times W)\times(H\times W)$, which means each position corresponds to a sub-attention map of size  $(H\times W)$. \cref{fig:model} illustrates such sub-attention maps for the same position in the non-local attribute attention and its reversed attention. The reversed sub-attention emphasizes the features neglected by the attribute sub-attention.
We approximate the reversed embeddings (or so called attribute reversal) after the reversed attention as the residuals after removing the learned attribute features from the original features.

\textbf{Non-local Channel Module (NCM).} While NSM extracts contextuality in the spatial view, we further propose to capture semantic contextuality from the channel view. The channel maps of high-level features can be viewed as response activation of specific classes. Therefore, establishing their interrelationships can explore semantic contextuality~\cite{DBLP:conf/cvpr/FuLT0BFL19,chen2017sca}. We employ NCM to extract the channel interdependencies. The structure and pipeline of NCM are similar to NSM, but with two-fold differences. First, we adopt Fully-Connected Networks (FCN) instead of CNN to generate the query, key, and value maps. The FCNs are designed using the idea of Squeeze-and-Excitation~\cite{hu2018squeeze} with the FC layers replacing the convolutional blocks. Second, the spatial module performs pooling at the last step, while the channel module performs pooling at first; thus, all embedding sizes during the process differ accordingly. Passing $x$ through NCM, we obtain the non-local channel embedding $e_{nc}$ and its reversal $e_{rnc}$ for the attribute and reversal-based object classification, respectively.

\textbf{Attribute classification.} The extracted non-local spatial and channel embeddings $e_{ns}$ and $e_{nc}$ are concatenated to form $e_{n}$ and fed to the attribute classifier $f_{ac}$ to predict the attributes. During training, we minimize the cross-entropy loss to improve the attribute compatibility: 
\begin{equation}
\resizebox{0.9\linewidth}{!}{$
    \mathcal{L}_{a}=\sum_{x,y=(a,o)\in\mathcal{S}}\mathcal{L}_{ce}(x,a)=-\sum_{x,y=(a,o)\in\mathcal{S}}\log f_{ac}(e_n,a)$}
\end{equation}
where $\mathcal{L}_{ce}$ denotes cross-entropy loss; $a$ is the ground-truth attribute label for $x$. $f_{ac}(e_{n},a)$ represents the probability of $a$, assigned by  $f_{ac}$ based on the input $e_{n}$.

\subsection{Local Networks for Objects}\label{sec:obj}

Existing works in CZSL often model object recognition as part of the composition task and treat it as equivalent of learning the attributes, thus ignoring how to better recognize objects from an object perspective.
We argue that the goal of object learning in CZSL is not only limited to transferring object knowledge in compositions, but also to improve object classification performance. A case for the latter comes from  
 related fields such as zero-shot image classification, where adopting the local attention mechanisms have led to successful attempts at extracting discriminative features~\cite{li2022balancing,guo2019progressive}, localizing distinct regions~\cite{ge2021semantic,li2022entropy}, etc. Thus we consider local attention for improved object learning.

\textbf{Local Spatial and Channel Module (LSM and LCM).} 
The structure of LSM is illustrated  in \cref{fig:model} (bottom center). A convolutional layer followed by the $\mathit{Sigmoid}$ function acts upon $z$ to produce the local attention weights and their reversed mappings (obtained by subtracting the weights from 1). We multiply $z$ with the two attention maps to obtain the local spatial embedding $e_{ls}$ and its reversal $e_{rls}$. Local and reverse-local channel embeddings $e_{lc}$ and $e_{rlc}$ are computed in a similar manner by LCM.

\textbf{Object classification.} To combine the local spatial and channel features, we concatenate $e_{ls}$ and $e_{lc}$ as $e_l$. We then use the object classifier $f_{oc}$ to predict objects supervised by the cross-entropy loss:
\begin{equation}
\small
 \mathcal{L}_{o}=\sum_{x,y=(a,o)\in\mathcal{S}}\mathcal{L}_{ce}(x,o)=-\sum_{x,y=(a,o)\in\mathcal{S}}\log f_{oc}(e_l,o)   
\end{equation}
where $o$ is the ground-truth object of $x$, and  $f_{oc}(e_{l},o)$ outputs the probabilities corresponding to the object labels.

\subsection{Attribute-Object Disentanglement}\label{sec:dis}

The non-local and local modules capture contextuality and locality for independent attribute and object recognition without considering their compositional nature. To account for the latter, we propose the reverse-and-distill strategy that disentangles the attribute and object features so that any unseen composition becomes perceptible.
As illustrated in \cref{fig:intro}, to disentangle the visual primitives, we regularize the attribute learning by the attribute- and object-reversals. The underlying reasoning for this is two-fold: 1) the object's feature map and its reversal are naturally disentangled; 2) if the attribute reversal contains much object information, the attribute features become less likely to contain object knowledge thus disentangled from the object features. Such attribute features are then entangled and largely overlapped with the object reversals due to the virtue of the first point. Note that these inferences also hold for object learning. 

\textbf{Reverse.} Owing to the aforementioned reasoning, we desire the object- and attribute-reversals to be sufficiently informative to predict attributes and objects, respectively. In this case, the attribute and object features would exclude information about each other thus, becoming disentangled. We combine non-local attribute-reversals $e_{rns}$ and $e_{rnc}$ into $e_{rn}$, and concatenate local object-reversals $e_{rls}$ and $e_{rlc}$ into $e_{rl}$. Then, $e_{rn}$ and $e_{rl}$ are swapped to be fed to the reversal-based object and attribute classifier, respectively. We guide the reverse learning with the cross-entropy loss:
\begin{equation}
\small
    \mathcal{L}_{r}=-(\sum_{x,y=(a,o)\in\mathcal{S}}\log f_{roc}(e_{rn},o)
    +\log f_{rac}(e_{rl},a))
\end{equation}

\textbf{Distill.} We also optimize the attribute features to learn from object reversal and the object features to learn from attribute reversal to enlarge the overlaps for further disentanglement. Intuitively, if the attribute features completely overlap with the object reversal, the attribute features would be disentangled from the object features due to the natural disentanglement between the object and its reversal. We introduce a knowledge distillation loss~\cite{hinton2015distilling} quantified by the Kullback–Leibler (KL) Divergence term to perform the teacher-student learning where the attribute- and object-reversals act as teachers:
\begin{equation}
\begin{aligned}
\small
    \mathcal{L}_{d}=&\sum_{x,y=(a,o)\in\mathcal{S}}\mathcal{KL}(f_{oc}(e_{l},o)\Vert f_{roc}(e_{rn},o))\\
    &+\mathcal{KL}(f_{ac}(e_{n},a)\Vert f_{rac}(e_{rl},a))
\end{aligned}
\end{equation}

\subsection{Training and Inference}

\textbf{Training objectives.} To enable collaborative learning of modules in DRANet, we define the overall training loss as:
\begin{equation}
\small
    \mathcal{L}_{czsl} =  \mathcal{L}_{a}+\mathcal{L}_{o}+\lambda_{1}\mathcal{L}_{r}+\lambda_{2}\mathcal{L}_{d}
\end{equation}
where $\lambda_{1}$ and $\lambda_{2}$ are hyper-parameters.

\textbf{Inference.} We fuse attribute and reversal-based attribute predictions, fuse object and reversal-based object predictions, and multiply the fusions to obtain final predictions:
\begin{equation}
\resizebox{0.9\linewidth}{!}{$
\begin{aligned}
    y' = &\argmax_{y=(a,o)\in Y^{\mathcal{T}}}((1-\eta_1)f_{ac}(e_{n},a)+\eta_{1}f_{rac}(e_{rl},a))\\
    &*((1-\eta_2)f_{oc}(e_{l},o)+\eta_{2}f_{roc}(e_{rn},o))
\end{aligned}$}
\end{equation}
where $\eta_1$ and $\eta_2$ modulate the fusion amounts of reversed classifier predictions.

\begin{table}[]
\small
\centering
\setlength{\tabcolsep}{1.5pt}
\begin{tabular}{lccc|cc|cccc}
\toprule
           &     &     &        & \multicolumn{2}{c|}{Training} & \multicolumn{4}{c}{Testing}                \\
Dataset    & a   & o   & p    & sp          & i            & sp & up & i   &  cw/p \\
\midrule
MIT-States & 115 & 245 & 28175  & 1262          & 30k          & 400  & 400    & 13k & 6\%                 \\
UT-Zappos  & 16  & 12  & 192    & 83            & 23k          & 18   & 18     & 3k  & 53\%                 \\
C-GQA      & 413 & 674 & 278362 & 5592          & 27k          & 888 & 923   & 5k  & 2\%     \\
\bottomrule
\end{tabular}
\caption{Datasets: a, o, p, i, sp, and up are the number of attributes, objects, all pairs, images, seen pairs, and unseen pairs. cw/p is the ratio of CW testing pairs to all pairs. }
\label{tab:datasets}
\end{table}

\section{Experiment}

\begin{table*}[]
\small
\centering
\begin{tabular}{l|cccc|cccc|cccc}
\toprule
\multirow{2}{*}{Method} & \multicolumn{4}{c|}{MIT-States}                             & \multicolumn{4}{c|}{UT-Zappos}                                & \multicolumn{4}{c}{C-GQA}                                   \\
                        & ~S~          & ~U~        & HM           & AUC          & ~S~          & ~U~        & HM            & AUC           & ~S~          & ~U~       & HM           & AUC           \\ \midrule
TMN~\cite{purushwalkam2019task}                    & 12.6          & 0.9           & 1.2          & 0.1          & 55.9          & 18.1          & 21.7          & 8.4           & NA            & NA           & NA           & NA            \\
AoP~\cite{nagarajan2018attributes}                    & 16.6          & 5.7           & 4.7          & 0.7          & 50.9          & 34.2          & 29.4          & 13.7          & NA            & NA           & NA           & NA            \\
LE+~\cite{misra2017red}                     & 14.2          & 2.5           & 2.7          & 0.3          & 60.4          & 36.5          & 30.5          & 16.3          & 19.2          & 0.7          & 1.0          & 0.08          \\
VisProd~\cite{misra2017red}                 & 20.9          & 5.8           & 5.6          & 0.7          & 54.6          & 42.8          & 36.9          & 19.7          & 24.8          & 1.7          & 2.8          & 0.33          \\
SymNet~\cite{li2020symmetry}                  & 21.4          & 7.0           & 5.8          & 0.8          & 53.3          & 44.6          & 34.5          & 18.5          & 26.7          & 2.2          & 3.3          & 0.43          \\
CGE$_{\text{ff}}$~\cite{naeem2021learning}                   & 29.6          & 4.0           & 4.9          & 0.7          & 58.8          & 46.5          & 38.0          & 21.5          & 28.3          & 1.3          & 2.2          & 0.30          \\
CGE~\cite{naeem2021learning}                     & \textbf{32.4} & 5.1           & 6.0          & 1.0          & 61.7          & 47.7          & 39.0          & 23.1          & \textbf{32.7} & 1.8          & 2.9          & 0.47          \\
CompCos$^{\text{CW}}$~\cite{mancini2021open}               & 25.3          & 5.5           & 5.9          & 0.9          & 59.8          & 45.6          & 36.3          & 20.8          & 28.0          & 1.0          & 1.6          & 0.20          \\
CompCos~\cite{mancini2021open}                 & 25.4          & \textbf{10.0} & \textbf{8.9} & \textbf{1.6} & 59.3          & 46.8          & 36.9          & 21.3          & 28.4          & 1.8          & 2.8          & 0.39          \\
VisProd$_{\text{ff}}$++~\cite{karthik2021revisiting}&24.6&6.7&6.6&1.0&58.3&47.1&39.3&22.8&27.2&2.1&3.3&0.46\\
VisProd++~\cite{karthik2021revisiting}&28.1&7.5&7.3&1.2&\textcolor{blue}{62.5}&51.5&41.8&\textcolor{blue}{26.5}&28.0&2.8&4.5&0.75\\
KG-SP$_{\text{ff}}$~\cite{karthik2022kg}                 & 23.4          & 7.0           & 6.7          & 1.0          & 58.0          & 47.2          & 39.1          & 22.9          & 26.6          & 2.1          & 3.4          & 0.44          \\
KG-SP~\cite{karthik2022kg}                   & 28.4          & 7.5           & 7.4          & 1.3          & 61.8          & \textcolor{blue}{52.1}          & \textcolor{blue}{42.3}          & \textcolor{blue}{26.5}          & \textcolor{blue}{31.5}          & 2.9          & 4.7          & \textcolor{blue}{0.78}          \\ \midrule
DRANet$_{\text{ff}}$           & 27.1          & 6.6           & 6.9          & 1.1          & 60.7          & 46.1          & 39.7          & 23.5          & 28.2          & \textcolor{blue}{3.1}          & \textcolor{blue}{5.0}          & 0.71          \\
DRANet          & 29.8          & \textcolor{blue}{7.8}           & \textcolor{blue}{7.9}          & \textcolor{blue}{1.5}          & \textbf{65.1} & \textbf{54.3} & \textbf{44.0} & \textbf{28.8} & 31.3          & \textbf{3.9} & \textbf{6.0} & \textbf{1.05}\\ \hdashline
\textemdash \textemdash Base Model              & 25.6   & 6.8     & 7.0  & 1.1  & 59.5  & 50.9    & 41.1 & 25.2 & 31.4 & 3.0    & 4.6 & 0.75 \\
\textemdash \textemdash ANet                                   & 28.9   & 7.2     & 7.4  & 1.3  & 61.0  & 53.7    & 42.9 & 27.3 & 30.6 & 3.5    & 5.4 & 0.88 \\
\textemdash \textemdash RANet            & \textcolor{blue}{30.9}   & 7.5     & 7.8  & 1.4  & 64.5  & 54.2    & 43.8 & 28.3 & 30.6 & 3.8    & 5.9 & 0.94 \\
 \bottomrule
\end{tabular}
\caption{\textbf{Main results} and the overall \textbf{module ablation}. The performance is evaluated by best accuracy on seen (S), unseen (U), their harmonic mean (H), and the area under the curve (AUC). ff represents fixing backbone during training. Best results are in bold. Second best results are in blue.}
\label{exp main}
\end{table*}

\subsection{Experiment Settings}

\textbf{Datasets and evaluation metrics.} We evaluate our model on three widely-used datasets: MIT-States~\cite{isola2015discovering} composing 115 attributes and 245 objects, UT-Zappos~\cite{yu2014fine,yu2017semantic} containing 16 attribute and 12 objects, and C-GQA~\cite{naeem2021learning} consisting of 413 attributes and 674 objects. 
We follow previous works~\cite{purushwalkam2019task,naeem2021learning} to split the datasets into seen and unseen compositions, and adopt the \textbf{Generalized CZSL}~\cite{naeem2021learning} setting where both seen and unseen pairs may appear at test time. The statistics of the split and datasets are shown in \cref{tab:datasets}. Note that unseen compositions are not revealed in OW-CZSL, \textit{i.e.,} the model may output non-existing pairs. For example, as shown in \cref{tab:datasets}, only 2\% out of all possible pairs occur in C-GQA test data.
We evaluate the model following the protocol in \cite{purushwalkam2019task,mancini2021open}: we calibrate a bias on seen compositions during testing and vary the bias to obtain the best seen accuracy (S), best unseen accuracy (U), best harmonic mean (HM) and the area under the curve (AUC).

\textbf{Implementation Details.} We follow prior practices~\cite{karthik2022kg,naeem2021learning} to adopt ResNet18~\cite{he2016deep} as our image encoder. Other modules in DRANet are built as one- or two-layer FCN or CNN. The model is trained end-to-end with Adam optimizer~\cite{kingma2014adam}. The learning rate is set to $5e-5$. 
The supplementary provides codes with detailed parameters.

\subsection{Comparisons with SOTAs}

We compare DRANet with approaches adapted from CW-CZSL~\cite{purushwalkam2019task,nagarajan2018attributes,misra2017red,li2020symmetry,naeem2021learning}, and methods designed for OW-CZSL~\cite{mancini2022learning,karthik2021revisiting,karthik2022kg}. Given the same data splits and evaluation protocols, we use the results reported in \cite{karthik2022kg} for competitors. 
Results are shown in \cref{exp main}. As can be seen, our DRANet achieves the best or comparable results on all datasets. In particular, DRANet yields 8.7\% and 34.6\% relative improvements of AUC over the second-best methods on UT-Zappos and C-GQA datasets, respectively. It also achieves impressive gains for the harmonic mean (HM) on the two datasets, i.e., 1.7\% and 1.3\%, respectively. HM is the key criterion among S, U, and HM, since it depicts the balance between both seen (S) and unseen classes (U). On MIT-State, our model performs the second-best inferior to CompCos~\cite{mancini2021open}. Although DRANet shows a lower HM with a gap of 1.0, the AUC gap drops to 0.1, indicating that the performance of our model is uniform and robust, albeit with a more modest peak compared with CompCos. 

A variant of our model that fixes the backbone during training (DRANet$_\text{ff}$) also performs the best among the fixed-backbone methods, demonstrating that our improvements are not derived from the image encoder.
The reasons for improvements are three-fold.
First, comparing methods containing two parallel attribute and object discriminators (DRANet, KG-SP, and VisProd++) with other methods that predict in the composition space, we find that for the OW-CZSL setting, modeling attributes and objects separately is more appropriate, and leads to better performance in general. Second, we propose the reverse-and-distill strategy to disentangle the attributes and objects, thus improving the generalization ability. Comparing KG-SP~\cite{karthik2022kg} with our model, both of which adopt two separate classification modules, our model shows superior performance on all criteria, proving that our models can transfer knowledge to unseen pairs better. Third, we adopt different non-local and local feature extractors designed based on distinct characteristics of attributes and objects, benefiting their recognition. Further analysis of the extractor structure is detailed in \cref{sec ablation}.

\begin{table}[]
\small
\centering
\setlength{\tabcolsep}{4pt}
\begin{tabular}{cl|cccccc}
\toprule
                              ~&        ~ & S & U & HM   & AUC  & HM-a & HM-o \\
\midrule
\multirow{4}*{\rotatebox{90}{\textbf{Attention}}} & Both Local                   & \textbf{62.6} & 52.0   & 42.3 & 26.8 & 52.7      & \textbf{73.9}   \\
~& Both Non-local                  & 62.5 & \textbf{53.7}   & 42.4 & 27.2 & \textbf{53.5}      & 73.4   \\
~&SwapA     &60.8	&52.0	&41.8 &26.3		&51.5	&72.6\\
~& ANet          & 61.0 & \textbf{53.7}   & \textbf{42.9} & \textbf{27.3} & 53.3      & 73.8   \\
\midrule
\multirow{3}*{\rotatebox{90}{\textbf{Reverse}}}                           & ANet w $\mathcal{L}_{r}$                                   & 64.1 & 53.0   & 42.6 & 27.7 & 53.1      & 72.8   \\
~& $a'*o'+a'_r*o'_r$                 & 64.1 & 53.7   & 43.2 & 28.1 & 53.1      & 73.0   \\
~& RANet            & \textbf{64.5} & \textbf{54.2}   & \textbf{43.8} & \textbf{28.3} & \textbf{53.5}      & \textbf{73.1}   \\
\midrule
\multirow{3}*{\rotatebox{90}{\textbf{Distill}}}                            & l-oriented      & 64.4 & 53.6   & 43.5 & 28.1 & 53.3      & 73.2   \\
~& n-oriented & 64.3 & 53.8   & 43.3 & 28.3 & 53.4      & 73.2   \\
~& DRANet &\textbf{65.1} & \textbf{54.3}   & \textbf{44.0} & \textbf{28.8} & \textbf{53.6}      & \textbf{73.5}  \\
\bottomrule
\end{tabular}
\caption{Detailed module design ablation. Best results are marked for each module.}
\label{exp modules}
\end{table}

\subsection{Ablation Study and Parameter Analysis}\label{sec ablation}

\textbf{Overall Module Ablation.} We compare DRANet with its three variants: Base Model without attentions and disentanglement, ANet adopting non-local and local attentions over the base model, and RANet that further equips reverse attention and reversal-based classification into ANet for disentanglement (without revers distillation compared to DRANet). Results are shown in \cref{exp main}. We find that HM and AUC increase with each additional module across all datasets, suggesting that 1) extracting attributes and objects with strengthened contextuality and locality is beneficial; 2) reverse classification and reverse distillation both improve the model's adaptability to unseen compositions. We then analyse the detailed design of each component in \cref{exp modules}.

\textbf{Design of Attentions.} We contrast ANet with adopting identical attention (both local or non-local) or swapped attentions (SwapA: non-local for objects and local for attributes) to extract attributes and objects . \cref{exp modules} shows that adopting non-local and local attention improves the attribute and object accuracy respectively, with ANet achieving the best HM and AUC while Swap performing the worst. This is consistent with our claim that attributes and objects are of different contextual dependencies and identical extractors may impair their discrimination.

\textbf{Incorporation of Reversal.} We analyze how incorporating reversal-based classification results can aid the final prediction. We namely compare only using reverse loss $\mathcal{L}_r$ for model optimization (ANet with $\mathcal{L} _r$), and two variants further incorporating reversal-based predictions in inference, \textit{i.e,} $a'*o'+a'_r*o'_r$ and $(a'+a'_r)*(o'+o'_r)$ (adopted by RANet ). As shown in \cref{exp modules},  integrating reverse learning helps improve the performance, with $(a'+a'_r)*(o'+o'_r)$ yielding a larger gain. $a'*o'+a'_r*o'_r$ and $(a'+a'_r)*(o'+o'_r)$ can be viewed as ensembles of two and four models, respectively (each product can be seen as the output of a distinct model). The performance gain is thus correlated with a better model ensemble that helps alleviate domain shift while increasing the robustness against noise~\cite{xu2019self}. 


\textbf{Orientation of Distillation.} We also evaluate the effect of distilling orientation in \cref{exp modules} by comparing DRANet with variants that: 1) treat attribute and reversal-based object classifier on top of non-local modules as teachers, namely n-oriented, 2) consider two classifiers built on local modules as teachers, namely l-oriented. We find that only DRANet aids further disentanglement on top of RANet. It may be because 1) DRANet performs mutual distillation between non-local and local modules, while the n- and l-oriented approaches rely on the local or non-local modules dominating the teacher-student learning, thus hurts the performance; 2) DRANet adopt reversals as teachers. Seeing~\cref{fig:intro} and comparing using reversals as teachers (Img2$\stackrel{teach}{\longrightarrow}$Img3$\stackrel{unravel}{\longrightarrow}$Img1) with as students (Img3$\stackrel{teach}{\longrightarrow}$Img2$\stackrel{reverse}{\longrightarrow}$Img1$\stackrel{unravel}{\longrightarrow}$Img3), the former is more straightforward, and the extra Img2$\stackrel{reverse}{\longrightarrow}$Img1 in the latter may cause gradient vanishing since reversing operation contains Sigmoid. Therefore it is better to use reversals as teachers instead of students.
\begin{figure}
\small
  \centering
  \begin{subfigure}{0.4\linewidth}
  \includegraphics[width=\textwidth]{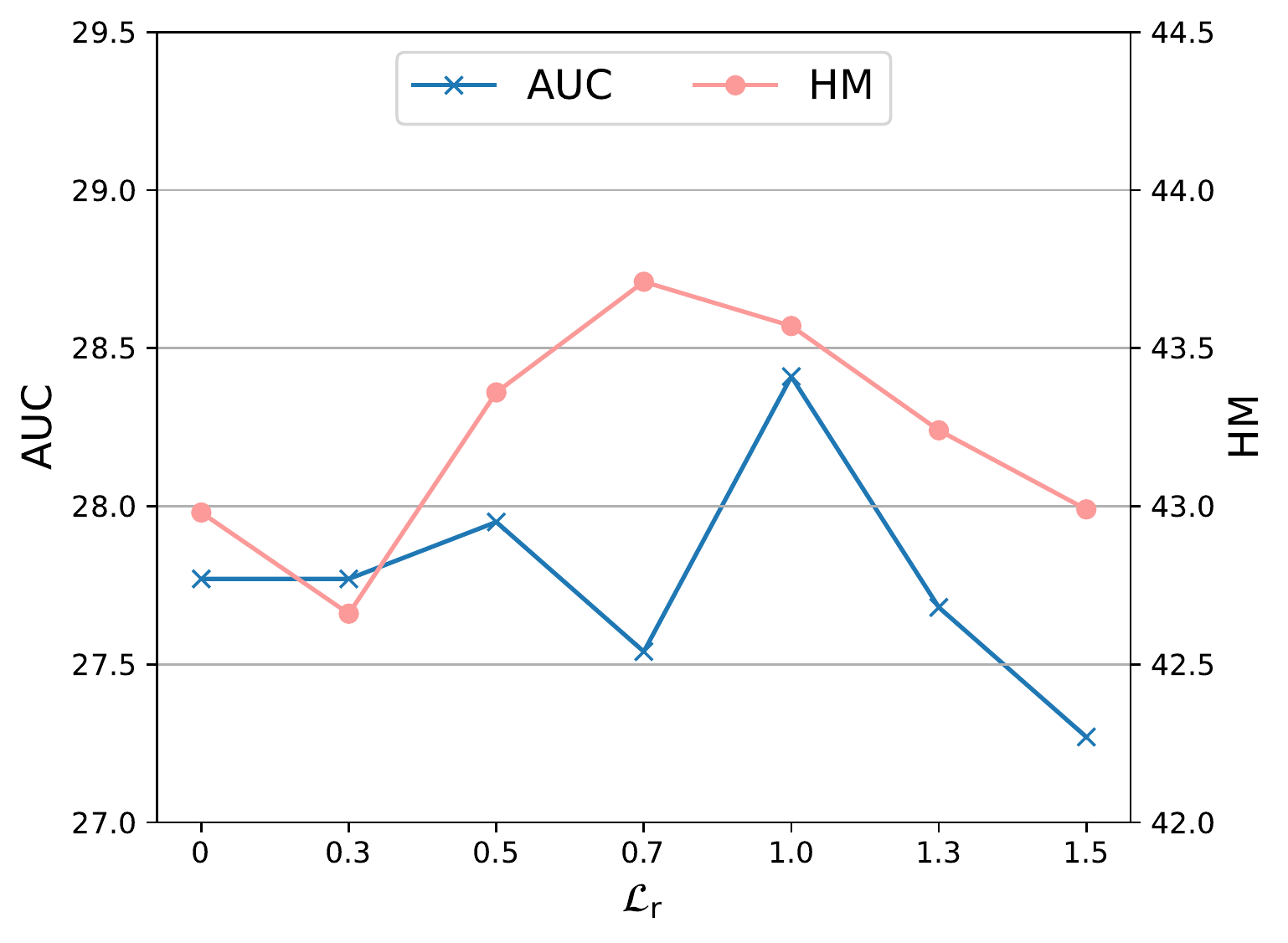}
    \caption{$\lambda_1$ of $\mathcal{L}_r$.}
    \label{fig:loss-b}
  \end{subfigure}
  \quad
  \centering
  \begin{subfigure}{0.4\linewidth}
  \includegraphics[width=\textwidth]{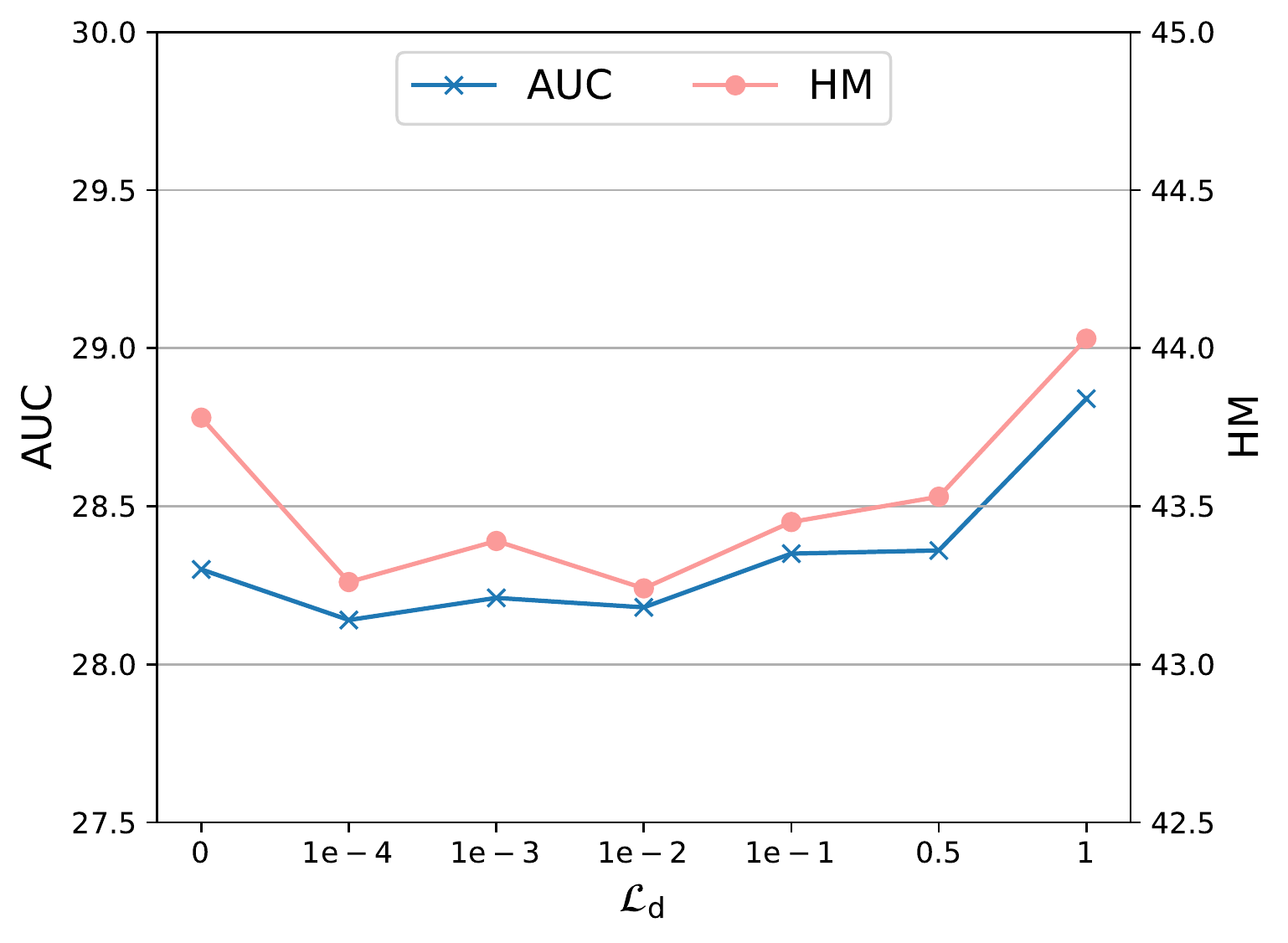}
    \caption{$\lambda_2$ of $\mathcal{L}_d$.}
    \label{fig:loss-a}
  \end{subfigure}
  \\
  \centering
  \begin{subfigure}{0.4\linewidth}
  \includegraphics[width=\textwidth]{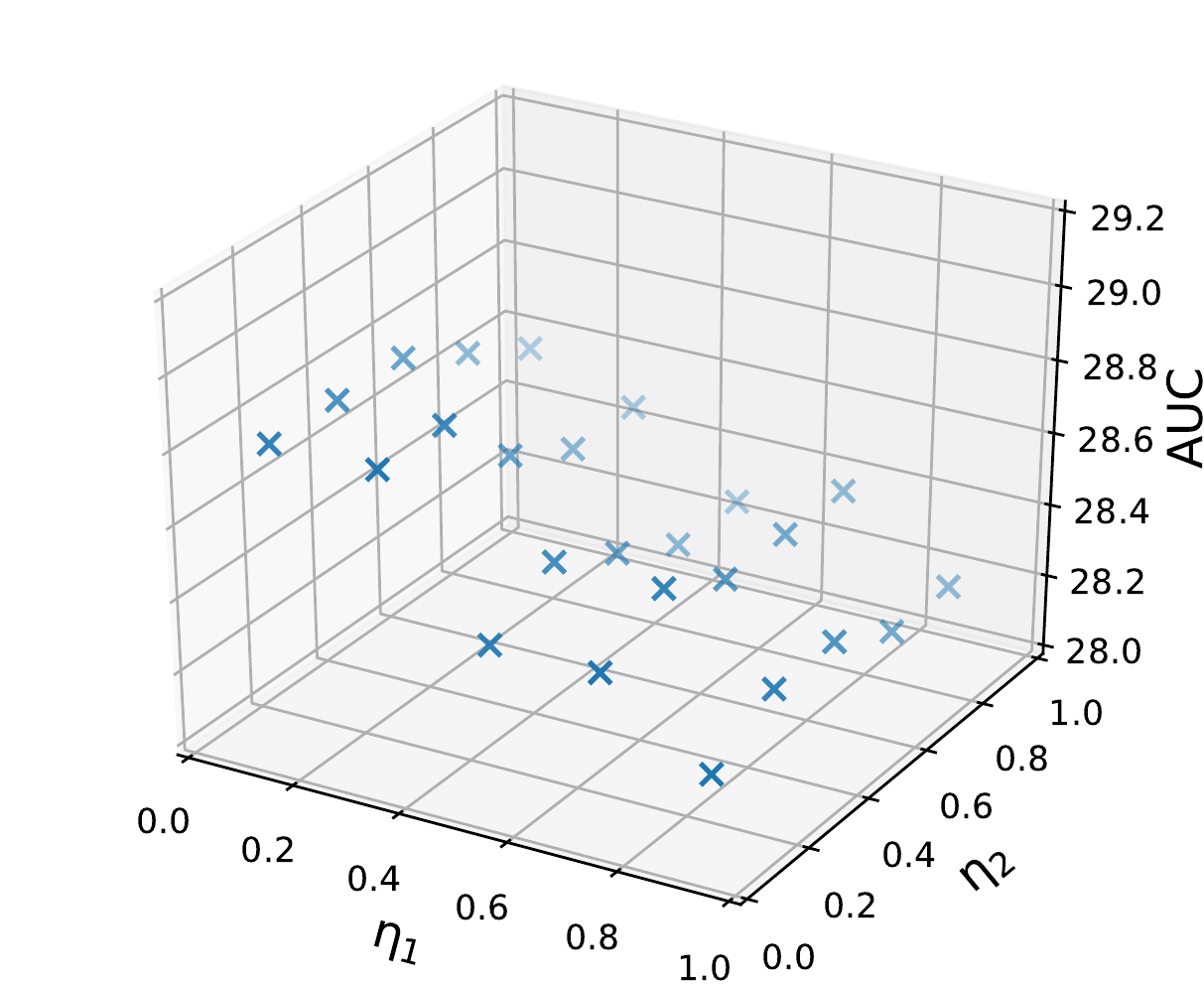}
    \caption{AUC.}
    \label{fig:fuse-b}
  \end{subfigure}
  \quad
    \centering
  \begin{subfigure}{0.4\linewidth}
  \includegraphics[width=\textwidth]{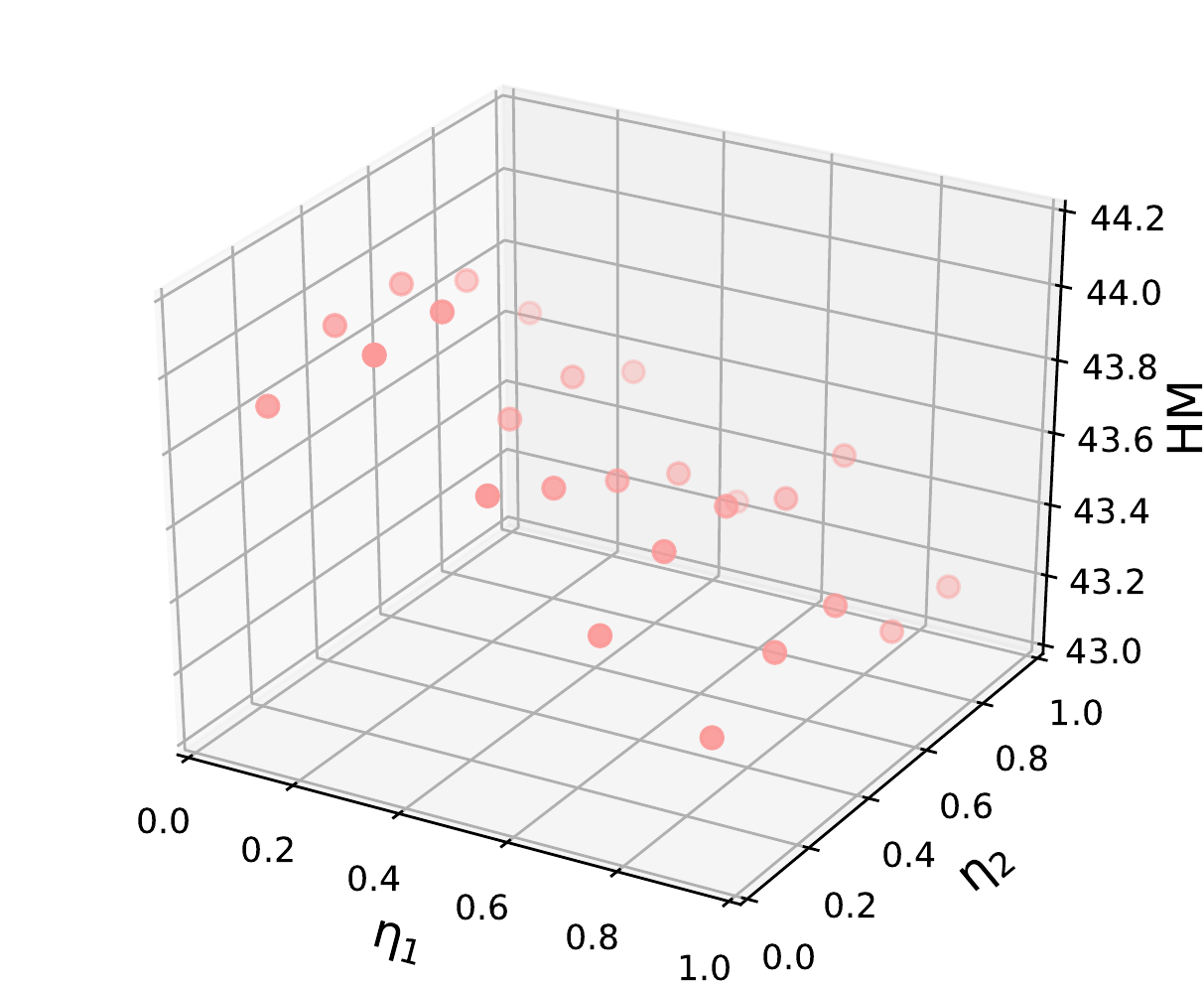}
    \caption{HM.}
    \label{fig:fuse-a}
  \end{subfigure}
  \caption{Loss and fusing ratios on UT-Zappos.}
  \label{fig:lossratio}
\end{figure}

\textbf{Hyper-parameter Analysis.} We also analyze model's sensitivity to hyper-parameters on UT-Zappos. \cref{fig:loss-b,fig:loss-a} show the results with varying loss ratios. We observe that on varying $\mathcal{L}_r$, the performance increases first and then decreases. This trend gets reversed while varying $\mathcal{L}_d$ with both the loss ratios achieving the best results around 1.0.
We also vary the fusion ratios ($\eta_1$, $\eta_2$) and show the results in \cref{fig:fuse-a,fig:fuse-b}. HM and AUC are best at (0.1, 0.3).


\begin{figure*}
\small
    \centering
    \begin{subfigure}{0.69\linewidth}
    \centering
    \includegraphics[width=0.92\textwidth]{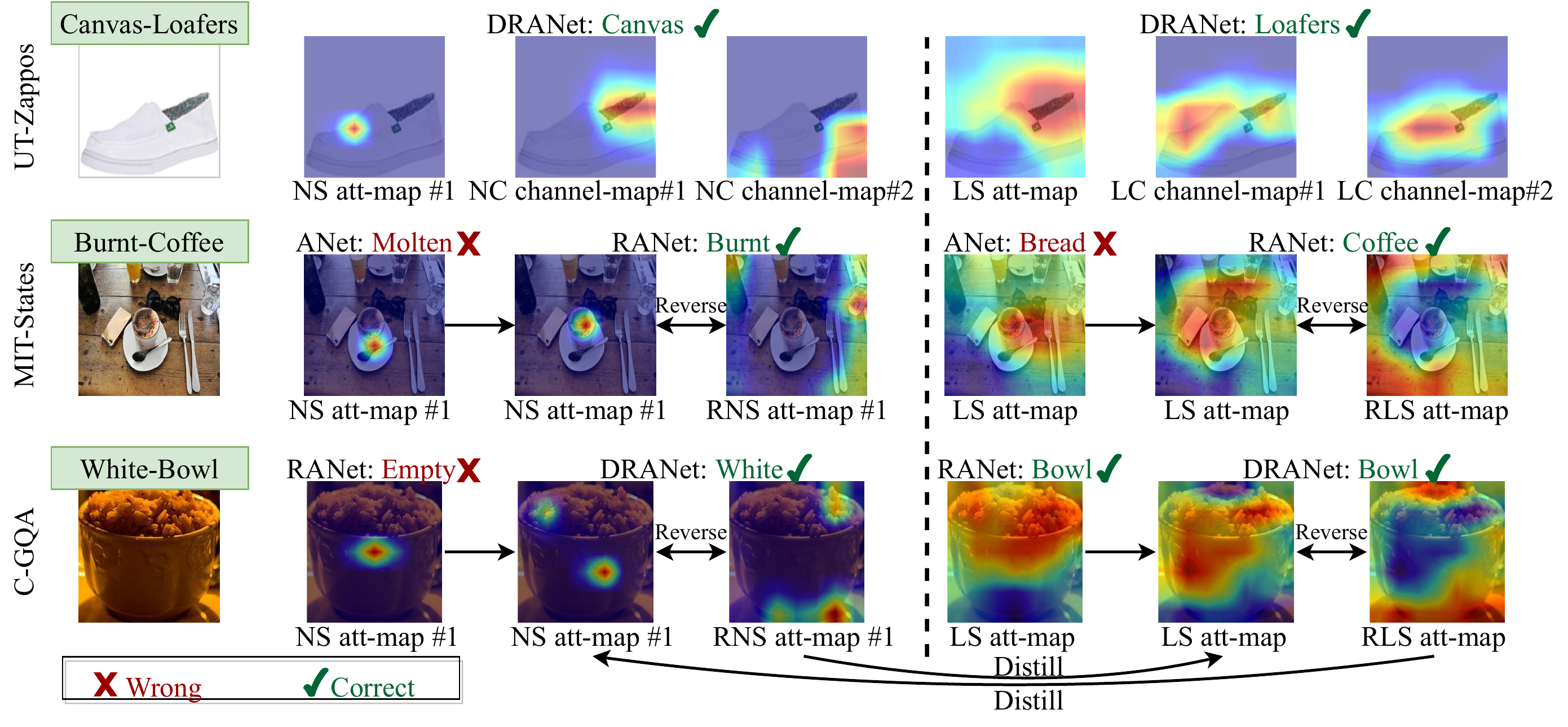}
    \caption{Attention and reverse-and-distill.}
    \label{fig:exp-att}
    \end{subfigure}
    \quad
    \begin{subfigure}{0.23\linewidth}
    \centering
    \includegraphics[width=0.89\textwidth]{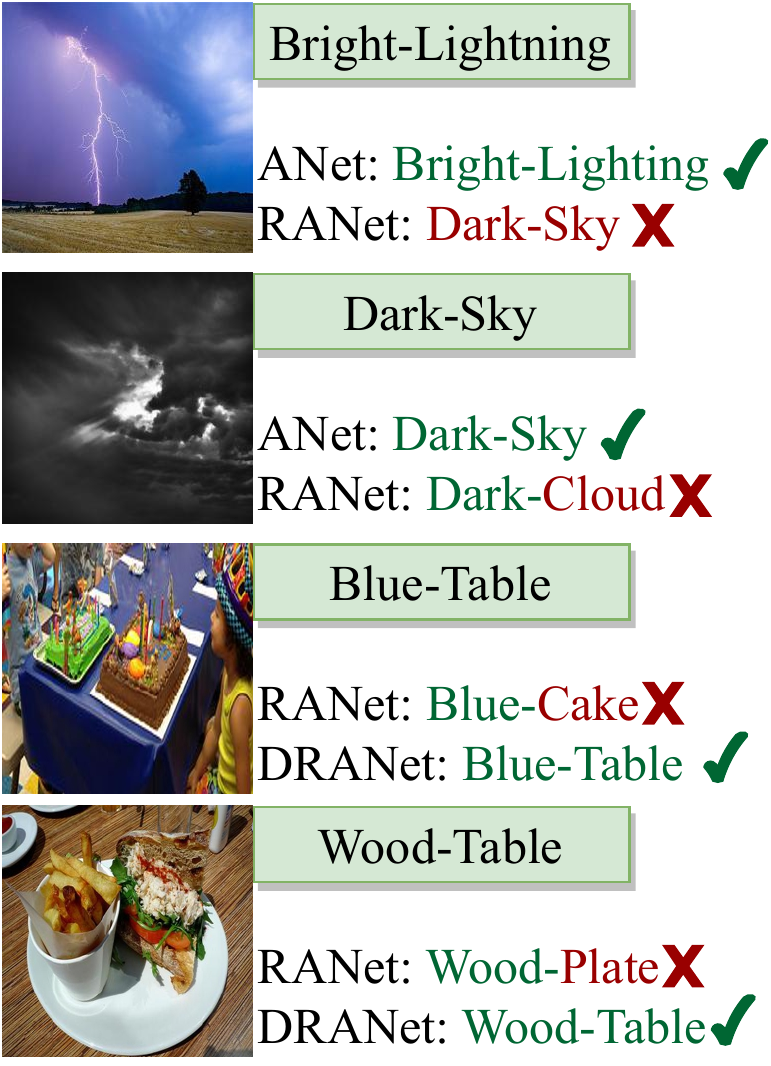}
    \caption{Limitation and extensibility.}
    \label{fig:exp-sample-2}
    \end{subfigure}
    \centering
    \\
\begin{subfigure}{0.82\linewidth}
\centering
    \includegraphics[width=0.92\textwidth]{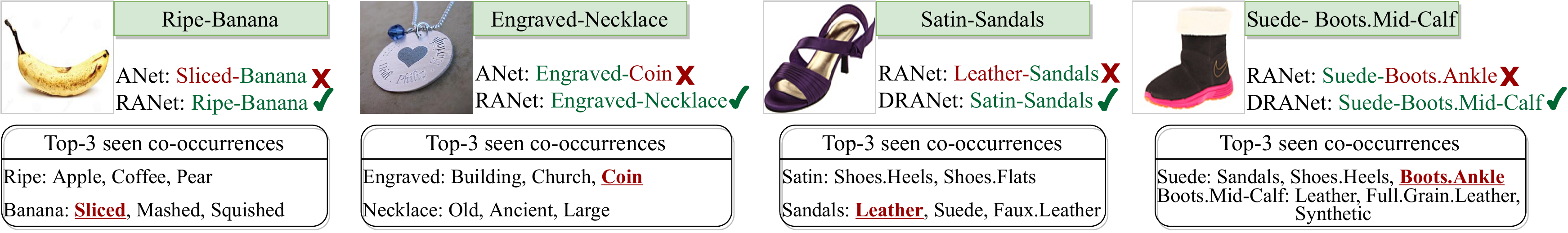}
    \caption{Disentanglement.}
    \label{fig:exp-visual}
\end{subfigure}
\caption{Visualization. (a) \textbf{Attention and reverse-and-distill.} For each image, the three activation maps on the left and the right  refer to the non-local (N) attention of attributes and the local (L) attention of objects, respectively. S, C and R further denote the spatial, channel and reverse attention maps while att-map and channel-map represent the spatial attention maps and channel feature maps, respectively. (b)-(c) \textbf{Qualitative results:} (b) Cases for limitations and extensibility of our model. (c) For unseen compositions, we show the top-3 frequent seen co-occurrences of their attributes and objects in training data, and the predictions of DRANet and its variants, to explore disentanglements.}
\label{fig:visuals}
\end{figure*}

\subsection{Visualization Results}

\textbf{Attentions and reverse-and-distill.} We choose samples from three datasets and visualize attention maps in \cref{fig:exp-att} to explain what attention learns and how the reverse-and-distill optimizes the attention. We visualize the local spatial attention directly, show the non-local attention corresponding to the pixel with the peak local attention weight, and display feature maps of some attended channels since it is hard to directly display channel attention. From image of Canvas-Loafers, we observe that the learned attention maps attend to discriminative regions. To identify Burnt-Coffee, we observe that ANet is fooled by the fork and knife to misclassify it as Molten-Bread, while RANet shifts its attention to the coffee and cup through the reversing strategy and thus predicts correctly. For White-Bowl, RANet ignores the rice and predicts it as Empty-Bowl, while the reverse attention distills the non-local attention to expand its focus from bowl to both rice and bowl thus producing the right label.

\textbf{Qualitative results.} We study the qualitative results to explore if visual disentangling is actually happening (\cref{fig:exp-visual}), and if happens, what are its limitations and extensibility (\cref{fig:exp-sample-2}). \textbf{Disentanglement:} As shown in \cref{fig:exp-visual}, we choose images of unseen compositions and display the top-3 frequent seen co-occurrences of ground-truth primitives. In the two leftmost images, ANet can be seen to predict correct attributes/objects but mispredict the images as seen compositions with the correct primitives due to the entanglement. For example, ANet recognizes Ripe-Banana as Sliced-Banana, where Sliced is the most frequent attribute co-occurring with Banana in training data. Similarly, ANet misclassifies Engraved-Necklace as Engraved-Coin. RANet enhances ANet with reverse attention to cut off co-occurrences; thus, it rectifies mistakes. Distilling further enlarges attribute-object gaps to  unravel features that RANet cannot handle. This is shown in the rightmost two images in \cref{fig:exp-visual}, where DRANet corrects entangled predictions of RANet to Satin-Sandal and Suede-Boots.Mid-Calf. 

\textbf{Limitations:} Reverse attention may 1) confuse the focal point of the image -- as shown in \cref{fig:exp-sample-2}, RANet identifies Bright-Lighting as Dark-sky and Dark-Sky as Dark-Cloud (although also correct); or 2) even lead to attribute-object inconsistency, \eg, misclassifying Blue-Table as Blue-Cake and Wood-Table as Wood-Plate when the images have cakes or plates on the table. The reason is that attention and reverse-attention reinforce attributes and objects independently. \textbf{Extensibility}: Limitation (1) inspires us to adopt reverse attention in multi-object recognition as it can find neglected information, such as dark sky around bright lightning.  Limitation (2) can be relieved by the distilling process as it coordinates attention and reverse-attention mutually (\eg, DRANet amends Blue-Cake to Blue-Table, and Wood-Table to Wood-Plate in \cref{fig:exp-sample-2}).


\section{Conclusion}

In this work, we propose a Distilled Reverse Attention Network (DRANet) to tackle the Open-World Compositional Zero-Shot Learning task. 
We capture attribute context-dependency and object local distinction through extractors tailored to their inherent discrepancies. We then design the reverse-and-distill strategy, which adopts reverse attention as the regularizer and the cross-distiller, to disentangle attribute and object features, thus better transferring recognition ability to unseen compositions.
Through comprehensive experiments, we prove the effectiveness of our model and achieve SOTA performance on three datasets. In addition, we highlight the limitations of our work, including entity inconsistency and focal confusion, which, however, may be beneficial for uncovering overlooked information, if extended to multi-object recognition in the future.

{\small
\bibliographystyle{ieee_fullname}
\bibliography{egbib}
}

\end{document}